\documentclass{article}
\usepackage{graphicx}
\usepackage{amsmath,amssymb,amsfonts}
\usepackage{lineno,hyperref}
\usepackage{algorithmic}
\usepackage{mathtools}
\usepackage{xcolor}
\usepackage{csvsimple}
\usepackage{multirow}
\usepackage{colortbl}
\usepackage{tabu}
\usepackage{booktabs}
\usepackage{tabularx}

\title{\textbf{Human and Machine Action Prediction Independent of Object Information}}

\author{Fatemeh Ziaeetabar$^{1*}$, Jennifer Pomp$^2$, Stefan Pfeiffer$^1$, Nadiya El-Sourani$^2$,\\ Ricarda I. Schubotz$^2$,
Minija Tamosiunaite$^{1,3}$ and Florentin W\"org\"otter$^{1}$}
\date{%
    $^1$Institute for Physics 3 - Biophysics and Bernstein Center for Computational Neuroscience (BCCN), University of G\"ottingen , G\"ottingen, Germany\\%
    $^2$Department of Psychology, University of M\"unster, M\"unster, Germany\\
    $^3$Department of Informatics, Vytautas Magnus University, Kaunas, Lithuania\\
    \textbf{Contacts:} fziaeetabar@gwdg.de (\textbf{corresponding author}), jennifer.pomp@uni-muenster.de, stefan.pfeiffer1@stud.uni-goettingen.de, nadiya.el-sourani @uni-muenster.de, rschubotz@uni-muenster.de, m.tamosiunaite@if.vdu.lt, worgott@gwdg.de\\
}
\begin{document}
\maketitle
\newpage
\textbf{Predicting other people's upcoming action is key to successful social interactions, enabling us to adjust our own behavior to the consequence of the others' actions well in advance. Previous studies on action recognition have focused on the importance of individual visual features of objects involved in an action as well as the action's context. Humans, however, can recognize actions performed with unknown objects or even when objects are only imagined (pantomime). Lack of recognizable visual object features must, thus, be compensated by other cues during action recognition. Therefore here we focus on the role of inter-object \textit{relations} that change during an action. To address this, we designed a virtual reality setup and tested recognition speed for ten different manipulation actions with fifty participants. All objects had been abstracted by emulating them with cubes such that participants could not infer an action using object information. Instead, participants had to rely only on the limited information that comes from the changes in the spatial relations that happen during an action between those cubes. In spite of these constraints, our results show that participants were able to predict actions in, on average, less than 64\% of the action's duration. Furthermore, we employed a computational model, the so-called enriched Semantic Event Chain (eSEC), which incorporates the information of spatial relations, specifically (a) objects' touching/untouching, (b) static spatial relations between objects and (c) dynamic spatial relations between objects in an action scene. After being trained by the same actions as those observed by our subjects, this model successfully predicted actions even better than humans. We show, using information theoretical analysis, that eSECs are able to make optimal use of individual cues, whereas humans seem to mostly rely on a mixed-cue strategy, which takes longer until recognition. Providing a better cognitive basis of action recognition may, on the one hand improve our understanding of related human pathologies and, on the other hand, also help in building robots for conflict-free human-robot cooperation. Our results may here open new avenues.}

\section*{}
\subsection*{\textbf{Action Recognition and Prediction in Humans}}
Human beings excel at recognizing actions performed by others, and they do so even before the action goal has been effectively achieved \cite{isik2017fast, wurm2012squeezing}. Thus, humans engage in action prediction. During this process, the brain activates a premotor-parietal network \cite{caspers2010ale} that largely overlaps with the networks needed for action execution and action imagery \cite{hardwick2018neural}. Though in recent years, some progress has been made towards computationally more concrete models of the mechanisms and processes underlying action recognition \cite{giese2015neural}, it still remains largely unresolved how the brain accomplishes this complex task.
Prediction of actions can rely on different sources of information. The present study focused on the fact that human observers exploit static and dynamic spatial relations between the objects in an action scene. Comparing manipulation of appropriate objects (i.e., normal actions) with manipulations of inappropriate objects (i.e., pantomime), we found that brain activity during action observation was largely explained by processing of the actor's movements \cite{schubotz2009case}. As a caveat, this finding may be explained by the particular movement-focused strategy subjects selected in this study where normal and pantomime actions were presented in intermixed succession. Other studies show that motion features are used by the brain to segment observed actions into meaningful segments and to update internal predictive models of the observed action \cite{schubotz2012fraction, kurby2008segmentation}. Correspondingly, individuals segment actions into consistent, meaningful chunks \cite{newtson1973attribution, newtson1976perceptual}, and intra-individually, they do so in a highly consistent manner, albeit high inter-individual variability \cite{schubotz2012fraction}. It has been argued that the objective quality of these chunks is that within the continuous sequence, breakpoints may convey a higher amount of information than the remainder of the event. Nevertheless, this suggestion remains speculative as long as we do not find a way to objectively quantify the flow of information that the continuous stream of input provides. This objectification is hampered by the fact that time-continuous information is highly variable with regard to spatial and temporal characteristics differing between action exemplars. Moreover, object information is a confounding factor in natural actions. As exemplars of object classes, individual objects provide information about possible types of manipulation the observer has learned these objects to be associated with \cite{schubotz2014objects, bach2014affordance, nicholson2017understanding}. For instance, knives are mostly used for cutting. Hence, objects can efficiently restrict the number of actions that an action observer expect to occur \cite{schubotz2014objects}. Speculatively, humans may use a mixed strategy exploiting object as well as static and dynamic spatial information, and this strategy may be adapted to current constraints; for instance, static and dynamic spatial information may become more relevant when objects are difficult to recognize.

In the present study, we tested the hypothesis that, in the absence of object and contextual (i.e., room, scene) information, action outcome prediction by human observers can be modeled as exploitation of static and dynamic spatial relations between objects involved in the action.

\subsection*{\textbf{Action Recognition and Prediction in Machines}}
Action recognition is a fundamental task in computer vision, which recognizes human actions based on the complete action execution in a video. In  other  words,  action  recognition  is  an  after-the-fact  video  analysis  task  that  focuses  on  the  present  state \cite{Kong_2018}. It has been studied for decades and is still a very popular topic due to its wide applications including human-computer interfaces \cite{Rautaray_2015}, visual surveillance \cite{Ger_2014}, video indexing \cite{Poleg_2016}, intelligent humanoids robots \cite{Zhang_2013}, ambient intelligence \cite{Ramos_2008} and more. The actions could be simple human actions in constrained situations \cite{Aggarwal_1999, Davis_1997, Moeslund_2001, Wallraven_2003} up to complex actions in cluttered scenes or in realistic videos \cite{Lan_2012, Laptev_2008, Patron_2012, Yao_2011, Gall_2013}.

Researchers have made great efforts to create an intelligent system that can recognize complex human actions in cluttered environments. But for a machine, an action in a video is just an array of pixels. Most of the time, the data of these pixels is mixed with noise, which must be eliminated during a pre-processing stage for example during pose estimation of moving humans \cite{Torras_2012}. It has no idea about how to convert these pixels into an effective representation, and how to infer human actions from that representation. These two problems are considered as action representation and action classification in action recognition, and many attempts \cite{Laptev_2005, Raptis_2013, Ji_2013, Carreira_2017} have been proposed to address these two problems \cite{Kong_2018}. Moreover, psychological studies on human behavior as well as reasoning \cite{Locke_1990} have pointed out the consequences of these two problems for both, understanding human cognition and intelligent system research. In this regard, there are methods that attempt to infer the type of action according to its (cognitive) consequences on the scene \cite{Yang_2013}.
The majority of the existing methods for human action recognition focus on low-level spatio-temporal features, which can be brittle, for example due to problems of intra class variability arising from different humans performing the same action \cite{Bulling_2014}. Approaches that use higher-level features \cite{Aksoy_2011, Ziaeetabar_2017, Ferm_2018} seem to be less affected by this. In this context, Ramirez-Amaro and coworkers have tried to consider human movements recognition from a semantic point of view  \cite{Karine_2019}.

Different from recognition, action \textit{prediction} is a before-the-fact video understanding task and is focusing on the future state. In many applications (e.g. in autonomous navigation, surveillance, health care and etc), intelligent machines do not have the opportunity to wait until an action has finished before making a reaction. Two examples can make this clear: 1) driver action prediction to prevent accidents or 2) prediction of a handicapped person's looming fall and proactive support by a robot. In these two examples post-hoc recognition will usually not help, but action prediction may prevent accidents.

For prediction, variability \cite{Din_2005} and incompleteness of the action execution \cite{Kong_2017} amplify the known problems in action recognition. After all, prediction is just ``recognition earlier in time''.
This topic is classified into one of three groups: 1) early action classification \cite{Ryoo_2011, Cao_2013}, 2) intention prediction \cite{Pei_2011, Li_2012} and 3) motion trajectory prediction \cite{Zhou_2011, Morrisand_2011}. While the first group has been extensively utilized to forecast short sequences of human motion, the second one remains elusive. Recently, Tanke et al. \cite{Gall_2019} approximated a person's intention by a symbolic representation and exploited it in conjunction with the observed human poses. These combined features are then used for predicting the ongoing action.

\par Manipulation prediction, which is the topic of this paper, can be understood as a sub-set within the above-discussed more general problem of human action prediction and mostly falls into groups 1 and 3. Ferm\"uller et al. have developed a recurrent neural network based method for manipulation action prediction \cite{Ferm_2018}. They depicted the hand movements before and after contact with the objects during the preparation and execution of actions and applied a method based on a recurrent neural network (RNN), where patches around the hand were used as inputs to the network. Moreover, there are some studies about hand motion trajectory recognition, which can be extended to prediction as well. They work in a causal way and can be also used for prediction. For example \cite{Elmezain_2009, Elmezain_2009_1} use a hidden Markov model-based continuous gesture recognition system utilizing hand motion trajectories and we have extended their methods from recognition to prediction in \cite{Ziaeetabar_2018}.

A central problem in all of these approaches is that action recognition (and prediction) heavily relies on time-continuous information (e.g. trajectories, movie sequences, etc.). This type of information, however, is highly variable, including spatial and temporal variations between action exemplars.

In an earlier study, we introduced a novel method in before-the-fact action recognition from incomplete action videos. We employed so-called extended semantic event chains (eSEC) \cite{Ziaeetabar_2018}, which are a strictly formalized and objective way to describe changes of static and dynamic spatial relation between objects involved in the scene. This approach allowed us to exactly determine discrete points in time at which combinations of spatial properties between objects in an action scene undergo significant change. Hence, this way we could disambiguate the ongoing action with regard to the intended goal. While the role of spatio-temporal changes along an action sequence had been considered important for action recognition before \cite{Pastra_2012, Yang_2014,Sum_2012,worgotter2020humans}, here we provide for the first time an analysis that highlights that humans employ a mixed strategy for accumulating evidence about the seen action. This is different from machine-vision, which can make use of the first unambiguous cue for recognition.

\section*{General Experimental Protocols and Methods}

\begin{figure}[!h]
    \centering
    \includegraphics[width=1\textwidth]{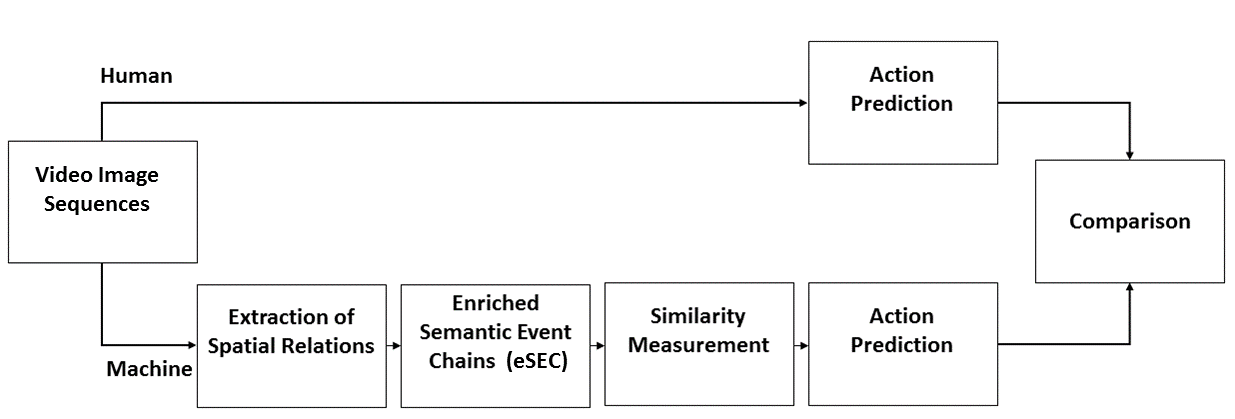}
    \caption{\small Experimental schedule}
    \label{steps}
\end{figure}

We designed a set of ten action scenarios in a virtual reality (VR) experiment to compare the predictive power of manipulation actions in humans and machine. Fig.~\ref{steps} illustrates the steps taken to this end. In the following we will briefly describe our experimental protocols; for details see Methods.

We employed ten actions: \textit{chop}, \textit{cut},\textit{ hide}, \textit{uncover}, \textit{put on top}, \textit{take down}, \textit{lay}, \textit{push}, \textit{shake}, and \textit{stir}. All objects, including hand and tools, were represented by cubes of variable size and color to serve object-agnostic (except the hand) action recognition. The hand was always shown as a red cube (Fig.~\ref{Exp}). Scene arrangements and object trajectories varied in order to generate a wide diversity in the samples of each manipulation action type. For each of the ten action types, 30 sample scenarios were recorded by human demonstration. All action scenes included different arrangements of several cubes (including distractor cubes) to ensure that videos were indistinguishable at the beginning.

\subsection*{Human Data Recording}
Forty-nine right-handed participants (20-68 yrs, mean 31.69 yrs, SD = 9.86, 14 female) took part in the experiment. One additional participant completed the experiments, but was excluded from further analyses due to an error rate of 14.7\%, classified as outlier. Prior to the testing, written informed consent was obtained from all participants.

Participants were given a detailed explanation regarding the stimuli and the task of the experiment. They were then familiarized with the VR system and shown how to deliver their responses during the experiment. The participants' task was to indicate as quickly as possible which action was currently presented.

\begin{figure}[!h]
    \centering
    \includegraphics[width=0.98\textwidth]{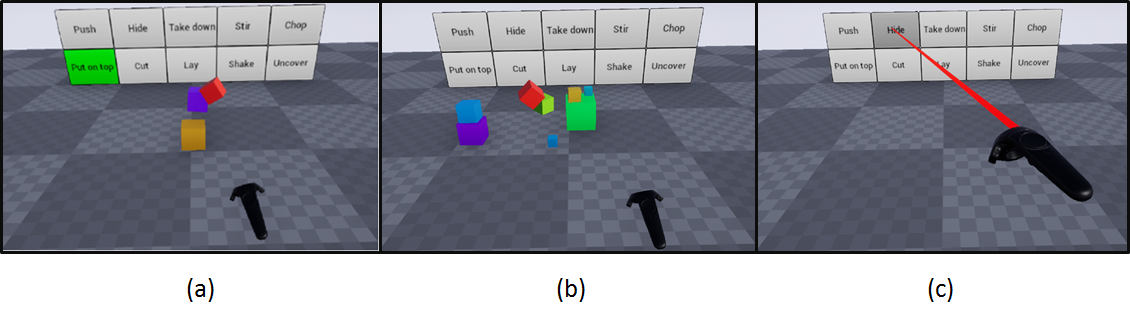}
    \caption{\small The VR experiment process, (a): experiment training stage: \textit{put on top} action, (b): experiment testing stage: action scene playing and (c): experiment testing stage: selecting the action type}
    \label{Exp}
\end{figure}

Every experiment started with a short training phase in which one example of each action was presented. During this demo version, the name of the currently presented action was highlighted in green on the background board (see Fig.\ref{Exp}~(a)). During the test stage, a total of $30\times 10$ action videos (trials) were shown to the participants where the red hand-cube entered the scene and performed an action (Fig.\ref{Exp}~(b)). When the action was recognized and the participant pressed the motion controller's button, the moment of this button press was recorded as response time. Concurrently, all cubes disappeared from the scene so that no post-decision cogitation about the action was possible. At the same time, the controller was marked with a red pointer added to its front. Hovering over the action of choice and pressing motion controller's button again recorded the actual choice and advanced the experiment to the next trial (Fig.\ref{Exp}~(c)). Participants were allowed to rest during the experiment, and continued the experiment after resting. Since participants mostly proceeded quickly to the next trial, the overall duration of the experimental session usually did not exceed one hour. All experimental data were analysed using different statistical methods (see Methods section).

\subsection*{Machine Action Prediction}
\label{SR}

The extended semantic event chain framework (eSEC) used for machine prediction of action makes use of object-object relations. We defined three types of spatial relations in our framework: 1)``Touching (T)'' and ``Non-touching'' (N) relations (TNR), 2) ``Static Spatial Relation'' (SSR) and 3)``Dynamic Spatial Relation'' (DSR).

Touching and non-touching relations between two objects were defined according to collision or non-collision between their representative cubes.

Static Spatial Relations (SSR) are the relative position of two objects in space. In this type of relations, no data from previous image sequences is needed and they can be determined in each image. These could, for example, be: `Above'' (\textbf{Ab}), ``Below'' (\textbf{Be}), ``Around'' (\textbf{Ar}), etc. For the complete list of SSRs see Methods.

\textit{Dynamic Spatial Relations, DSR} are the relations between objects\textsc{\char13} movements (when either or both of them move or are fixed). Here, different from SSR, some information from the previous frames (e.g., distance-related parameters) between each pair of objects is needed for DSR. Some examples are: ``Moving Together'' (\textbf{MT}), ``Halting Together'' (\textbf{HT}), etc. (for complete list, see Methods).

Importantly, eSEC do not make use of any \textit{real} object information. Objects remain abstracted (like in the VR experiments). We defined five abstract object types that play an essential role in any manipulation action and call them the \textbf{fundamental objects} (see Table \ref{obj_def}). Fundamental objects 1, 2, and 3 \textit{obtain} their role in the course of an action: they are numbered according to the order by which they encounter transitions between the relations N (non-touching) and T (touching). For example, `fundamental object ``1''' obtains its role given by 'number 1' by being the first that encounters a change in touching (usually this is the object first touched by the hand).

Note that not all fundamental objects defined in Table \ref{obj_def} are always existing in a specific action. Only \textit{hand}, \textit{ground} and \textit{fundamental object 1} are necessarily present in all analyzed actions. The action-driven ``birth'' of objects 1, 2, and 3 automatically leads to the fact that irrelevant (distractor) objects are always ignored by the eSEC analysis.

Thus, the maximal number of relations that had to be analyzed for an action was set by defined relations between fundamental objects: Given five object roles, there were $C(5,2)=10$ possible combinations leading to ten relations for each type (N/T, SSR, DSR), resulting in 30 relations in total.
\begin{table}[!h]
\caption{Definition of the fundamental objects during a manipulation action \cite{Ziaeetabar_2018}.}
\label{obj_def}
\begin{center}
\scalebox{0.8}{
\begin{tabular}{|c|c|c|}
\hline

\textbf{Object} & \textbf{Definition} & \textbf{Remarks}\\
\hline
\textbf{Hand} & The object that & Not touching anything at\\
 & performs an action. & the beginning and at the end of\\
 & & the action. It touches at least\\
 & & one object during an action.\\
\hline
\textbf{Ground} & The object that supports & It is extracted as a ground\\
 & all other objects except & plane in a visual scene.\\
 & the hand in the scene. &\\
 \hline
\textbf{1} & The object that is the \textbf{first}&Trivially, the first transition \\
 &to obtain a change&will always be a touch \\
 &in its T/N relations.&by the hand.\\
\hline
\textbf{2} &The object that is the \textbf{second}&Either T$\rightarrow$N or N$\rightarrow$T relational\\
 &to obtain a change&change can happen.\\
 &in its T/N relations.&\\
\hline
\textbf{3} &The object that is the \textbf{third}&Either T$\rightarrow$N or N$\rightarrow$T relational\\
 &to obtain a change&change can happen.\\
 &in its T/N relations.&\\
\hline
\end{tabular}}
\end{center}
\end{table}

\subsubsection*{The eSEC) Matrix as an Action Descriptor for Machine Prediction}
\label{eSEC_section}
\begin{figure}[!h]
    \centering
    \includegraphics[width=0.85\textwidth]{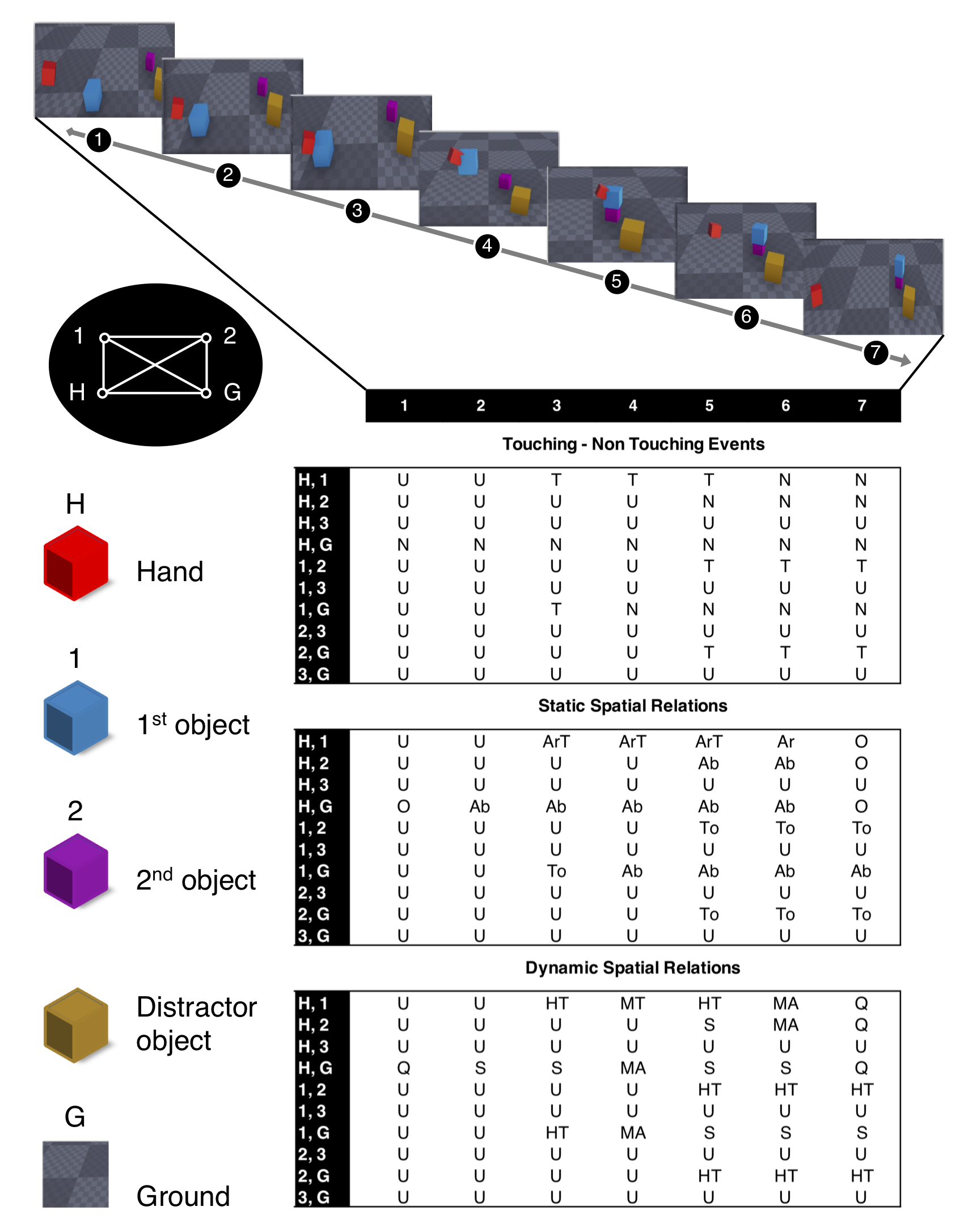}
    \caption{\small Description of a ``Put on top'' action in the eSEC framework with possible relation graph between all objects. Only hand and ground are pre-specified, object 1 is the one first touched by the hand, object 2 the next where a touching/un-touching (T/N) change happens and object 3 in this case remains undefined (U) in all rows as there are no more T/N changes. This leads to the graph on the top left that shows all possible relations. Abbreviations in the eSEC are: U: undefined, T: touching, N: non-touching, O: very far (static), Q: very far (dynamic), Ab: above, To: top, Ar: around, ArT: around with touch, S: stable, HT: halt together, MT: move together, MA: moving apart, GC: getting close. Note that the two leftmost columns are identical for all actions as they indicate the starting situation before any action. The top, middle and bottom ten rows of the matrix indicate TNR, SSR and DSR between each pair of fundamental objects in a ``put on top'' action, respectively. }
    \label{eSEC_Put}
\end{figure}

The Enriched Semantic Event Chain (eSEC) framework is inspired by the original Semantic Event Chain (SEC) approach \cite{Aksoy_2011}. The original SEC checks only touching (T) and non-touching (N) relations between each pair of fundamental objects in all frames of a manipulation scene and focuses on transitions (changes) in these relations. In the eSEC, the wealth of relations (see Methods section, Fig.~\ref{SRDR} for all of them) are embedded into a matrix-form representation, showing how the set of spatial relations changes throughout the action \cite{Ziaeetabar_2018}. Fig.\ref{eSEC_Put} shows the eSEC matrix for a ``put on top'' action and demonstrates how the set of all the different relations changes throughout this action.

\subsection*{Measuring Predictive Power}
\label{pred_section}
Importantly, since action types differed in duration, we assessed predictive performance not in absolute time, but relative to the length of each video. Using a probabilistic approach, we assessed at which time point any action would be predictable by its eSEC. To this end, we divided our data (eSEC tables) into \textit{train} and \textit{test} samples and performed a column-by-column comparison. That is, similarity values between the eSECs were derived by comparing each test action's eSEC to the every member of the training sample. We defined an action as ``predicted'' when the average similarity for one class remained high, while similarity for all other classes was low in this column. The similarity measurement algorithm between two eSEC matrices is explained in the Method section.

The eSEC column at which prediction happens is called ``prediction column''. That way, \textit{event-based predictive power} is defined as:
\begin{equation}
P=(1-\frac{column(\alpha)}{Total(\alpha)})*100\%
\end{equation}
where $column(\alpha)$ is the ``prediction column''  and $Total(\alpha)$ is the total number of columns in the action $\alpha$ eSEC table.

The \textit{event-based predictive power} provides the ratio of the event number we have predicted to the total number of events. Obviously, the lower the values of this measure, the lower the predictive power.


\section*{Results}

In the human reaction time experiments, response times that exceeded the length of the action video were treated as time-outs and corresponding trials (13 out of 14700) were excluded from further analyses.

Regarding learning effects, hence, possible trends in performance change along an experiment, correlation analyses showed a very small significant reduction effect in error rates $(rs = -.05, p < .05, n = 1470, one-sided)$ and a small significant deterioration effect for human predictive power $(rs = -.21, p = <.001, n = 1470, two-sided)$.

\begin{figure}[!h]
    \centering
    \includegraphics[width=1\textwidth]{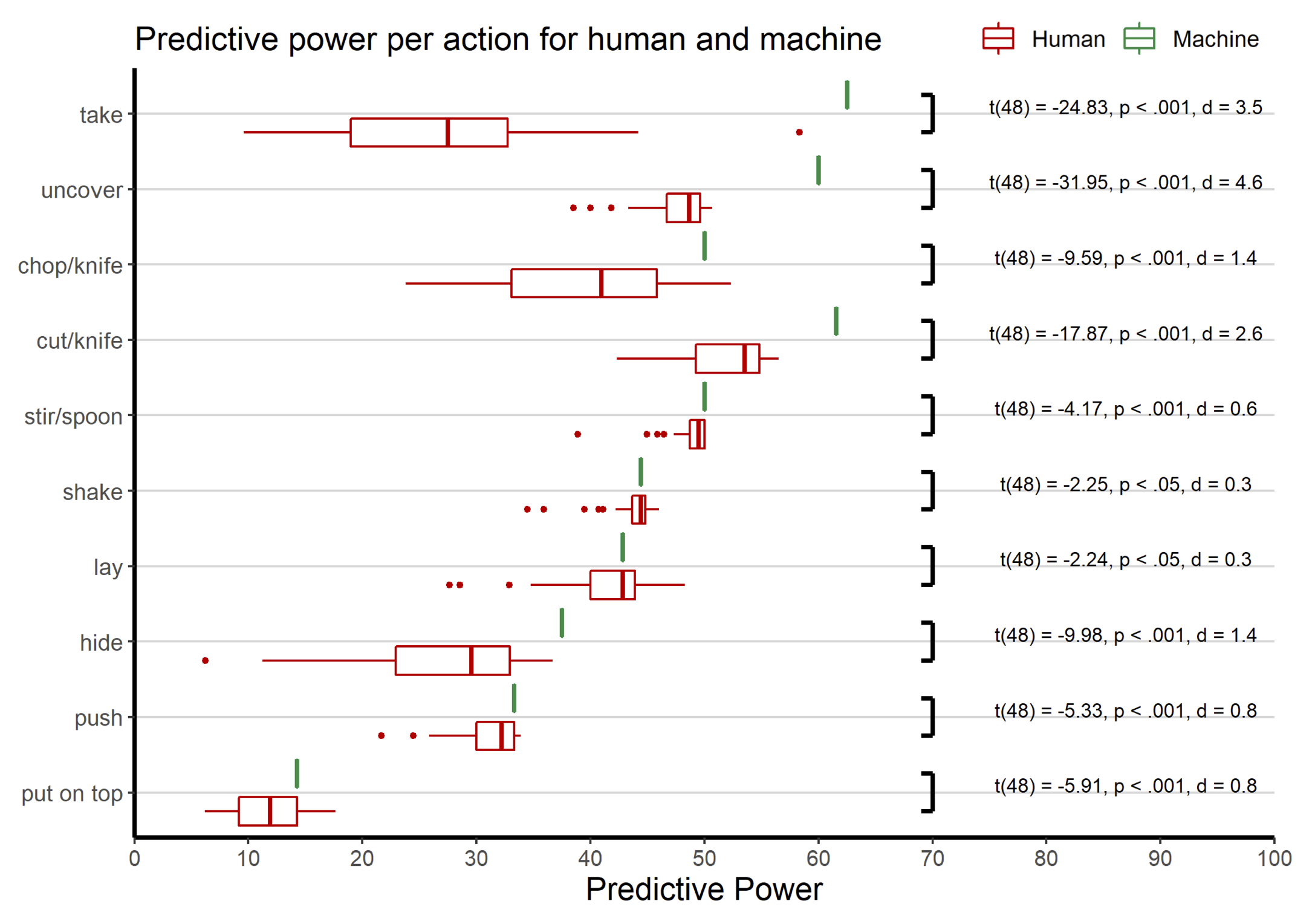}
    \caption{\small Mean predictive power of human and machine. t-values and p-values according to the t-tests per action.}
    \label{model_comp_Jenny_final}
\end{figure}

Human predictive power was analyzed using a repeated measures ANOVA with action as within-subjects factor. Mauchly’s test indicated that the assumption of sphericity was violated
$(\chi^{2}(44) = 310.45, p < .001)$, therefore degrees of freedom were corrected using Greenhouse-Geisser estimates of sphericity $(\epsilon= 0.37)$. The main effect of action was significant $(F(3.29,157.74) = 434.85, p < .001, {\eta_{p}}^2 = .901)$. As shown in Fig.\ref{model_comp_Jenny_final}, predictive power varied strongly between actions. For instance, put and take actions were not correctly classified before most columns of the video (88\% and 72\%, respectively) were already presented, whereas cut, stir and uncover did only need about half (48\%, 51\%, and 52\%, respectively) of the video time.

Separate one-sample t-tests per action for human vs. machine predictive power consistently showed higher predictive power for the algorithm $(ts < -2, ps < .05)$. See details in Fig.~\ref{model_comp_Jenny_final}. Predictive power ranged from 14.3\% to 62.5\% for the machine, whereas humans predictive power ranged from 6.2\% to 58.3\%. On average, the machine spared observation of the remaining 45.6\% of the video columns, humans the remaining 37\%. In half of the actions (\textit{take, uncover, hide, push} and \textit{put on top}), this difference reached a very large effect size $(ds > 1)$. Interestingly, most pronounced differences not in terms of effect size but in terms of overall sampling time emerged for actions that were most quickly classified by the algorithm (take, uncover, cut). For take actions, humans sampled twice as many columns (72\%) as the optimal performing algorithm (38\%).

\begin{figure}[!h]
    \centering
    \includegraphics[width=1\textwidth]{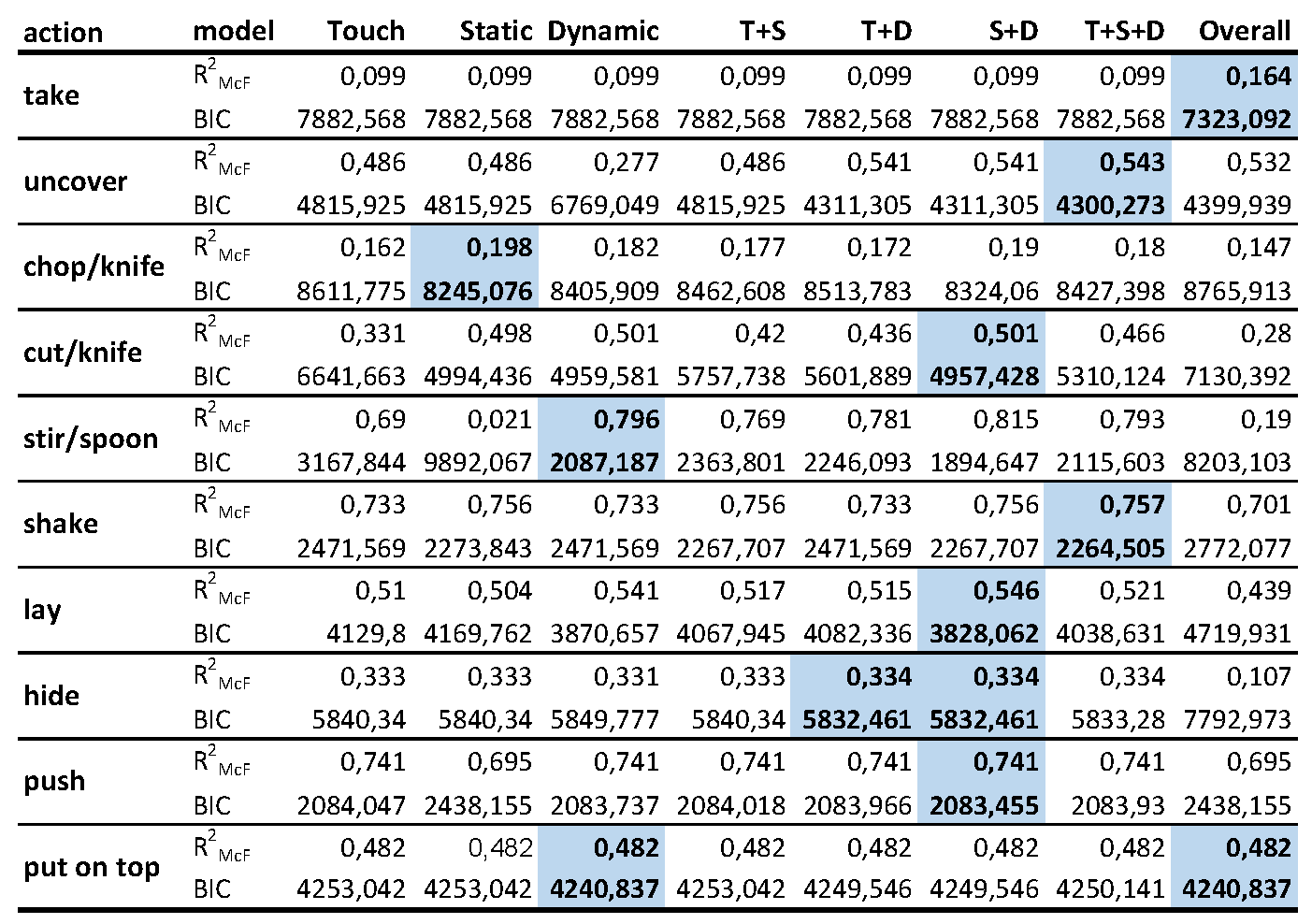}
    \caption{\small Fitting different models to the actions. (Abbreviations are shortened to allow to encode combinations by a short ``+'' annotation. We have Touch=T=TNR, Static=S=SSR, Dynamic=D=DSR. This leads to different combinations: T+S, T+D, S+D, T+S+D, where ``Overall'' refers to treating all eSEC columns independently of their individual information contents (see Methods).}
    \label{models}
\end{figure}

\begin{figure}[!h]
    \centering
    \includegraphics[width=1\textwidth]{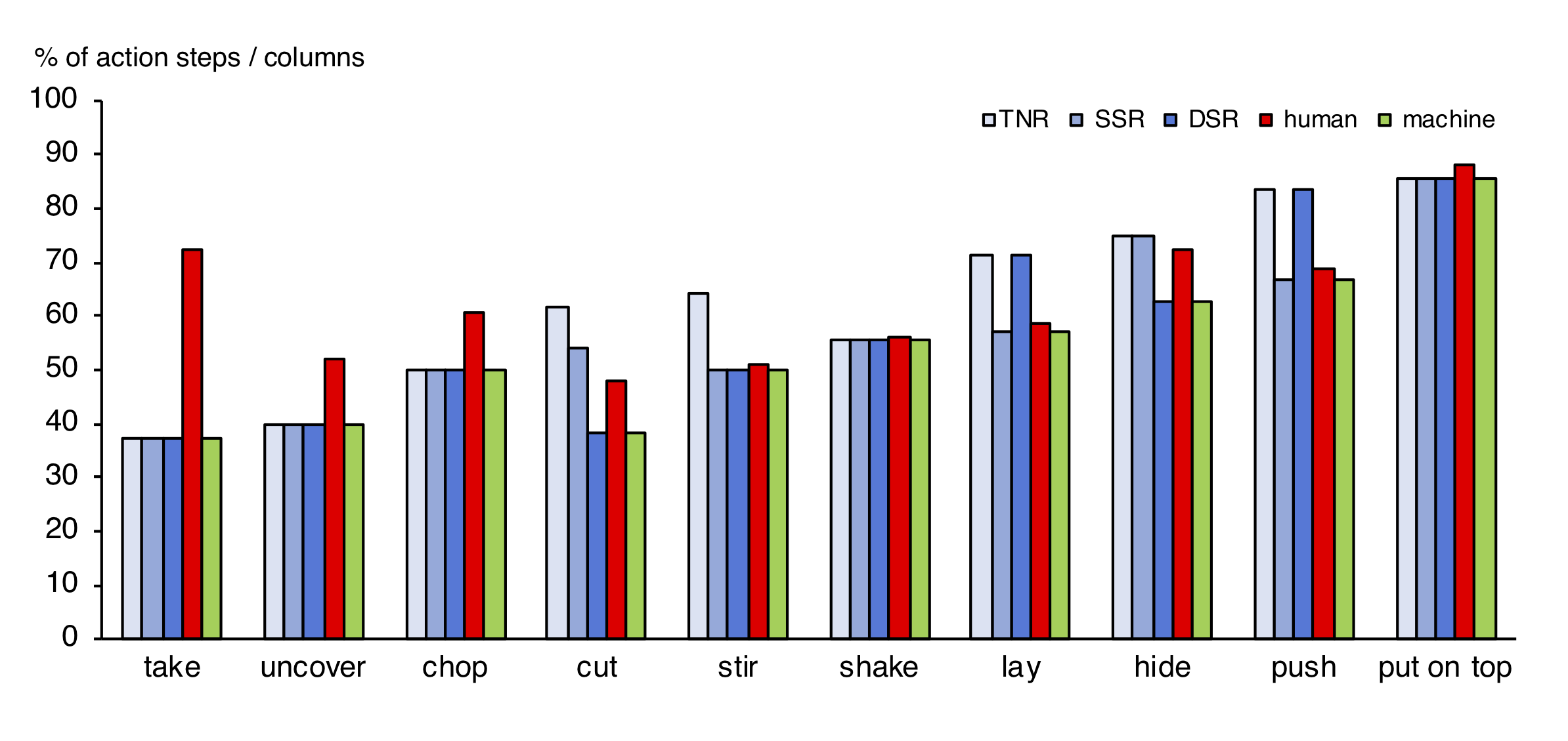}
    \caption{\small Comparison of human (red bars) and machine (green bars) predictive performance. Blue bars indicate the relative amount (percentage) of action steps elapsed per action, before the TNR (light blue), DSR (blue) or SSR (dark blue) model provided maximal local informational gain, enabling a secure prediction of the respective action. For instance, the 5th eSEC-column of the overall 13 eSEC-columns describing the \textit{cut} action provided a unique description in terms of DSR. That is, after around 38\% of these action's columns, the \textit{cut} action could be predicted on the basis of DSR information, and this is what the algorithm did, as indicated by the green bar of equal length. In contrast, humans correctly predicted the \textit{cut} action at the 6th (mean 6.26) column, corresponding to 48\% of this action, exploiting both dynamic and static spatial information (cf. Figure \ref{models} for this outcome).}
    \label{strategy_final}
\end{figure}

Logistic regressions revealed significant results for the eight models for each action respectively. All models tested significantly against their null model $(ps < .001)$. Figure \ref{models} shows McFadden $R^2$ and BIC per model per action. Shaded cells indicate which model fits best human action prediction behavior based on the BIC. Deploying the AIC (Akaike-Information-Criterion) yielded similar results.

As to the type of information exploited for prediction, we found marked differences between human and machine strategies. The machine behavior was perfectly predicted by the biggest local gain in information, i.e., by transition into the column where the action code became unique for the respective action (Fig.~\ref{strategy_final}). For instance, when dynamic information was the first to provide perfect disambiguation between competing action models, the algorithm always followed this cue immediately (this was the case for \textit{cut} and \textit{hide}). Likewise, static information ruled machine behavior for \textit{push} and \textit{lay}, reflecting the earliest possible point of certain prediction in these actions. Human suboptimal behavior was nicely reflected by the fact (see Fig.~\ref{models}) that for \textit{cut} and \textit{hide}, subjects considered a combination of both dynamic and static spatial information (where they should have focused on dynamic information); the same strategy was applied to \textit{push} and \textit{lay}, where subjects should have better followed static information only.

Notably, when all three types of information were equally beneficial (\textit{take, uncover, shake,} and \textit{put}), human performance was best modeled by a combination of all three types of information, with the exception of \textit{chop}, where subjects followed static spatial information. A post-hoc paired-sample t-test showed a significant effect of \textit{informational difference} $(t(48) = 15.95, p <.001, d_{z} = 2.3)$. The z-transformed difference between mean human and machine predictive power was explicitly larger for informationally indifferent actions (M = 2.1) than for informationally different actions (M = 1.1). Expressed in non-transformed values, humans showed 12\% less predictive power than the algorithm for informational indifferent action categories, but only 5\% for the informational different ones.

\subsection*{Discussion}
With the comprehensive entry of robots into various aspects of human life, it is necessary to pay more attention to establish appropriate interactions between humans and robots. These interactions must take place in a cognitive context that is ``understandable'' to both humans and machines. To improve on this, recently, Andriella et al. \cite{Torras_2020} have proposed a cognitive system framework for brain-training exercise based on human robot interaction. One of the central challenges for a proper interaction is the need to identify action of the human and make those ``understandable'' for the machine. This can be achieved by equipping the robot with an action encoding that allows action recognition and prediction.

Humans predict actions based on different sources of information, but we know only little about how flexible these sources can be exploited in case that others are noisy or unavailable. In the present study, we modeled human action prediction by an algorithm that was solely based on spatial information in terms of touching and untouching events between objects, their static spatial and dynamic spatial relations. This so-called enriched semantic event chain (eSEC) framework, which has been derived from older ``grammatical'' approaches towards action encoding \cite{Aksoy_2011, Pastra_2012, Yang_2014, worgotter2020humans, Aksoy_2017} has significant predictive power, because each eSEC column captures highly indicative changes in the spatial and temporal relationships between objects and hand in an action. Thus, as soon as columns are different from one case to another, actions can be discriminated and become, thus, predictable. In an older study \cite{Ziaeetabar_2018}, we compared the predictive power of the eSEC framework with a Hidden Markov Model (HMM) algorithm, which represents the current state of the art in computer vision-based action recognition \cite{Elmezain_2009, Elmezain_2009_1}. This was done on two real data-sets and we found that the average predictive power of eSECs is slightly above 60\% while $P<35$\% for the HMM-based approach.

In the present study, we tested how optimal human action prediction is when only spatial information is available. To this end we used action videos which were highly abstracted dynamic displays containing cubic place holders for all objects including hands, so that any information about real-world objects, environment, context, situation or actor were completely eliminated.

For such kind of action videos, we found that the algorithm was significantly faster than humans for predicting actions, i.e., for assigning the ongoing video to one out of ten basic action categories, before the video was completed. This difference was significant for each single action category. On average, humans achieved about 91\% of the predictive power of the machine. Based on an information theoretic approach, further analyses revealed that humans did not select the optimal strategy to disambiguate actions as fast as possible: While the machine reliably detected the earliest occurrence of disambiguation between the ongoing action and all other action categories, as indicated by the highest gain in information at the respective action step, human subjects did so in only half of the action categories. Instead, humans unswervingly applied a mixing strategy, concurrently relying on both dynamic and spatial information in 8 out of 10 action types.

This strategy was particularly disadvantageous for actions that were equally well predictable based on either static (N/T, SSR) or dynamic (DSR) information. Particularly in these - one may say - \emph{informationally indifferent} cases, humans were significantly biased towards prolonged decisions: here, they showed 12\% less predictive power than the algorithm as compared to 5\% for the informationally different actions.

In sum, the human bias towards using mixing strategies, combining static and dynamic spatial information, and to prolong decisions for informational indifferent action categories establish overall poorer human predictive power. In principal, these two effects may result from the same general heuristics of human action observers, to always exploiting several sources of information rather than relying on the first available source only. In other words, individuals seem to prioritize correct over fast classification of observed actions.

Our study was restricted to a number of ten possible actions, whereas in everyday life, the number of potentially observable actions is much higher, resulting in higher uncertainty and higher competition among these potential actions. Speculatively, the human bias to employ mixing exploitation strategies may be better adapted to disambiguate actions among this broader range of action classes. Future studies have to enlarge the sample of concurrently investigated actions to test this assumption and to increase overall ecological validity.


\subsection*{Limitations}

We controlled for a number of additional sources of information that humans are known to exploit in natural actions. Especially, object information provides an efficient restriction on to-be-expected manipulations \cite{schubotz2014objects, Ruddle_2002, Gupta_2007, Hrkac_2015, El-Sourani_2018, El-Sourani_2019}. It remains to be tested how non-spatial object information potentially interacts with the exploitation of static and dynamic spatial relations between objects involved in actions. Moreover, actions occur in certain contexts and environments that further restrict the observer's expectation, for instance with regard of certain classes of actions \cite{wurm2012squeezing, Shapovalova_2011, Zheng_2012, Wurm_2017, Wurm_COM_2012}.

Furthermore, our approach did not take into account all dynamic and static spatial information provided by human action. For instance, we restricted dynamic spatial information to between-object change, whereas in natural action, we would also register dynamic within-hand change. Thus, actors shape their hands to fit the to-be-grasped object already when starting to reach out for it \cite{Ingram_2010, Jean_1995}, providing a valuable pointer to potentially upcoming manipulations and goals \cite{Heumer_2008}. Likewise, gaze information plays a role in natural action observation \cite{Land_2009}, as the actors' looking to an object draws the observer's attention to the same object \cite{Fathi_2012}, and hence, potentially upcoming targets of the action.

\subsection*{Concluding remarks}

Describing action with a grammatical structure  \cite{Pastra_2012, Sum_2012, Yang_2014} such as eSEC \cite{Ziaeetabar_2017, Ziaeetabar_2018}, renders a simple and fast framework for recognition and prediction in the presence of unknown objects and noise. This robustness lends itself to an intriguing hypothesis, which is asking to what degree such an event-based framework might help young infants to bootstrap action knowledge in view of the vast number of objects that they have never encountered before. In terms of spatial relations (as implemented in the current eSEC framework), the complexity of an action is far smaller than the complexity of the realm of objects with which an action can be performed, even when only considering a typical baby's environment. Clearly, this approach has proven to be beneficial for robotic applications \cite{Aein_2019} and we plan to extend it to complex actions and interactions between several agents to examine the exploitation and exploration of predictive information during cooperation and competition.


\section*{Detailed Methods}



\subsection*{Virtual Reality System}
\label{VRS_section}
The main components of our VR system include computing power (for 3D data processing), head mounted display (for showing the VR content) and motion controllers (as the input devices).
A Vive VR headset and motion controller released by HTC in April 2016 with a resolution of 1080 x 1200 per eye, have been used as our VR system.  The ``roomscale'' system, which provides a precise 3D motion tracking between two infrared base stations, is the main advantage of this headset, which creates the opportunity to record and review actions for experiments on a larger scale of up to 5 meters diagonally.
The Unreal Engine 4 (UE4) is a high performance game engine developed by Epic Games and is chosen as the game engine basis of this project. It has built-in support for VR environments and the Vives motion controllers.

\subsection*{Scenario Recording}

In order to make VR-movies for the 10 different actions, 30 variants of each action were recorded by two members of BCCN team (a 23 year old undergraduate male and a 30 year old doctoral student female). They implemented a VR platform by using C++ code structure. The motion controller is the core input component of the VR environment and they provided a separate function for each button on that by C++ programming. The designed system included three different modes. First, a mode to record new actions for the experiment; second, a mode to review in, and last, the experiment itself. To keep the controls as simple as possible and to avoid a second motion controller without implementing a complex physics system, the recording mode was split into two sub-modes: A single-cube recording mode (for single, mostly static cubes) and a two-cubes recording mode (for object manipulation).


\subsection*{Stimuli}
\label{VRE_section}


Actions were defined as follows:\\

\textit{Chop:} The hand-object (hereafter: hand) touches an object (tool), picks up the object from the ground, puts it on another object (target) and starts chopping. When the target object has been divided into two parts, the tool object untouches the pieces of the target object. After that, the hand puts the tool object on the ground, untouches it, and leaves the scene.\\
\textit{Chop} scenarios had a \textbf{mean length} of \textbf{17.86} s (\textbf{SD = 3.56, range = 13-27}).\\\\
\textit{Cut:} The hand touches an object (tool), picks up the object from the ground, puts it on another object (target) and starts cutting. When the target object was divided into two parts, the tool object untouches the pieces of the target object. After that, the hand puts the tool object on the ground, untouches it, and leaves the scene.\\
\textit{Cut} scenarios had a \textbf{mean length} of \textbf{19.50} s (\textbf{SD = 3.13, range = 13-25}).\\\\
\textit{Hide:} The hand touches an object (tool), picks up the object from the ground, puts it on another object (target) and starts coming down on the target object until it covers that object thoroughly. Then the hand untouches the tool object and leaves the scene.\\
\textit{Hide} scenarios had a \textbf{mean length} of \textbf{13.43} s (\textbf{SD = 2.40, range = 9-20}).\\\\
\textit{Uncover:} The hand touches an object (tool), picks up the object from the ground. The second object (target) emerges as the tool object is raised from the ground, because the tool object had hidden the target object. After that, the hand puts the tool object on the ground, untouches it, and leaves the scene.\\
\textit{Uncover} scenarios had a \textbf{mean length} of \textbf{12.66} s (\textbf{SD = 3.20, range = 9-21}).\\\\
\textit{Put on top:} The hand touches an object, picks up the object from the ground and puts it on another object.  After that, the hand untouches the first object and leaves the scene.\\
\textit{Put on top} scenarios had a \textbf{mean length} of \textbf{10.90} s (\textbf{SD = 2.006, range = 8-16}).\\\\
\textit{Take down:}  The hand touches an object that is on another object, picks up the first object from the second object and puts it on the ground. After that, the hand untouches the first object and leaves the scene.\\
\textit{Take down} scenarios had a \textbf{mean length} of \textbf{10.60} s (\textbf{SD = 3.04, range = 6-18}).\\\\
\textit{Lay:} The hand touches an object on the ground and changes its direction (lays it down) while it remains touching the ground. After that, the hand untouches the object and leaves the scene.\\
\textit{Lay scenarios} had a \textbf{mean length} of \textbf{11.23} s (\textbf{SD = 1.79, range = 8-15}).\\\\
\textit{Push:} The hand touches an object on the ground and starts pushing it on the ground. After that, the hand untouches the object and leaves the scene.\\
\textit{Push} scenarios had a \textbf{mean length} of \textbf{12.56} s (\textbf{SD = 1.73, range = 9-17}).\\\\
\textit{Shake:} The hand touches an object, picks up the object from the ground and starts shaking it. Then, the hand puts it back on the ground, untouches it and leaves the scene.\\
\textit{Shake} scenarios had a \textbf{mean length} of \textbf{12.10} s (\textbf{SD = 2.05, range = 9-17}).\\\\
\textit{Stir:} The hand touches an object (tool), picks up the object from the ground, puts it on another object (target) and starts stirring. After that, the hand puts the tool object on the ground, untouches it, and leaves the scene.\\
\textit{Stir} scenarios had a \textbf{mean length} of \textbf{20.23} s (\textbf{SD = 4.67, range = 14-31}).\\\\

\subsection*{Machine Action Prediction}
The methods described in the following are largely identical to those used in \cite{Ziaeetabar_2018} and \cite{worgotter2020humans}.

\subsubsection*{Spatial Relations}
The details on how to calculate static and dynamic spatial relations are provided below. Here we start first with a general description.

1) Touching and non-touching relations (TNR) between two objects were defined according to collision or non-collision between their representative cubes.

\begin{figure}[!h]
    \centering
    \includegraphics[width=0.67\textwidth]{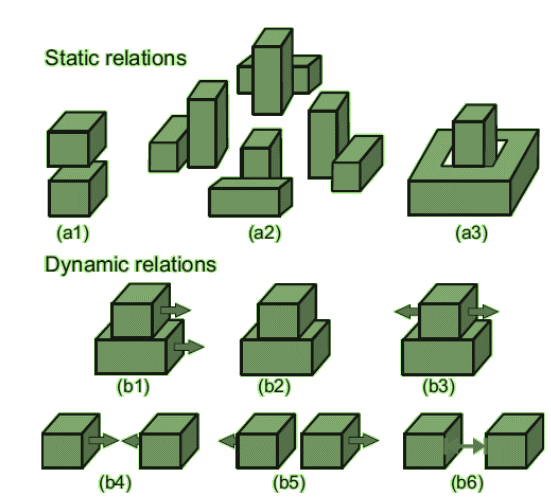}
    \caption{\small (a) Static Spatial Relations: (a1) Above/Below, (a2) Around, (a3) Inside/Surround.
(b) Dynamic Spatial Relations: (b1) Moving Together, (b2) Halting Together, (b3) Fixed-Moving Together, (b4) Getting Close, (b5) Moving Apart, (b6) Stable.}
    \label{SRDR}
\end{figure}

2) Static spatial relations (SSR) included : `Above'' (\textbf{Ab}), ``Below'' (\textbf{Be}), ``Right'' (\textbf{R}), ``Left'' (\textbf{L}), ``Front'' (\textbf{F}), ``Back'' (\textbf{Ba}), ``Inside'' (\textbf{In}), ``Surround'' (\textbf{Sa}). Since ``Right'', ``Left'', ``Front'' and ``Back'' depend on the viewpoint and directions of the camera axes, we combined them into ``Around'' (\textbf{Ar}) and used it at times when one object was surrounded by another. Moreover, ``Above'' (\textbf{Ab}), ``Below'' (\textbf{Be}) and ``Around'' (\textbf{Ar}) relations in combination with ``Touching'' were converted to ``Top'' (\textbf{To}), ``Bottom'' (\textbf{Bo}) and ``Touching Around'' (\textbf{ArT}), respectively, which corresponded to the same cases with physical contact. Fig.~\ref{SRDR}~(a1-a3) shows static spatial relations between two objects cubes. If two objects were far from each other or did not have any of the above-mentioned relations, their static relation was considered as Null (\textbf{O}). This led to a set of nine static relations in the eSECs: \textbf{{SSR}} = \{Ab, Be, Ar, Top, Bottom, ArT, In, Sa, O\}. The additional relations, mentioned above: \textbf{R}, \textbf{L}, \textbf{F}, \textbf{Ba} are only used to define the relation Ar=around, because the former four relations are not view-point invariant.

3) Dynamic Spatial Relations (DSR) require to make use of the frame history in the video. We used a history of 0.5 seconds, which is an estimate for the time that a human hand takes to change the relations between objects in manipulation actions. DSRs included the following relations: ``Moving Together'' (\textbf{MT}), ``Halting Together'' (\textbf{HT}), ``Fixed-Moving Together'' (\textbf{FMT}), ``Getting Close'' (\textbf{GC}), ``Moving Apart'' (\textbf{MA}) and ``Stable'' (\textbf{S}). DSRs between two objects cubes are shown in Fig.~\ref{SRDR}~(b1-b6). MT, HT and FMT denote situations when two objects are touching each other while: both of them are moving in a same direction (MT), are motionless (HT), or when one object is fixed and does not move while the other one is moving on or across it (FMT). Case \textbf{S} denotes that any distance-change between objects remained below a defined threshold of $\xi=1$ cm during the entire action. All these dynamic relations cases are clarified in Fig.~\ref{SRDR}~(b). In addition, \textbf{Q} is used as a dynamic relation between two objects when their distance exceeded the defined threshold $\xi$ or if they did not have any of the above-defined dynamic relations. Therefore, dynamic relations make a set of seven members: \textbf{DSR} = \{MT, HT, FMT, GC, MA, S, Q\}.

Finally, whenever one object became ``Absent'' or hidden during an action, the symbol \textbf{(A)} was used for annotating this condition. In addition, we use the symbol \textbf{(X)} whenever one object was destroyed or lost its primary shape (e.g. in \textit{cut} or \textit{chop} actions).

\subsubsection*{Object Types}
\label{OTS}
An exhaustive description of the five fundamental object types had been given in the main text and shall not be repeated here.

\subsection*{Mathematical Definition of the Spatial Relations}
\label{SR_app}
\begin{figure}[!h]
    \centering
    \includegraphics[width=1\textwidth]{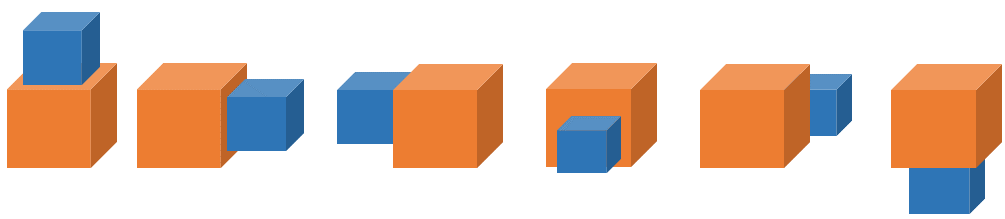}
    \caption{\small Possible situations that two cubes touch each other}
    \label{TN}
\end{figure}

As mentioned above, touching and non-touching relations between two objects are defined according to collision or non-collision between their representative cubes. 3D collision detection is a challenging topic which has been addressed in \cite{Torras_2001}. But, because the objects in our study are just cubes, we interpreted the contact of one of the six surfaces of one cube with one of the other cube's surfaces (see Fig.\ref{TN}) as touching event and this can be detected easily.

\begin{figure}[!h]
    \centering
    \includegraphics[width=1\textwidth]{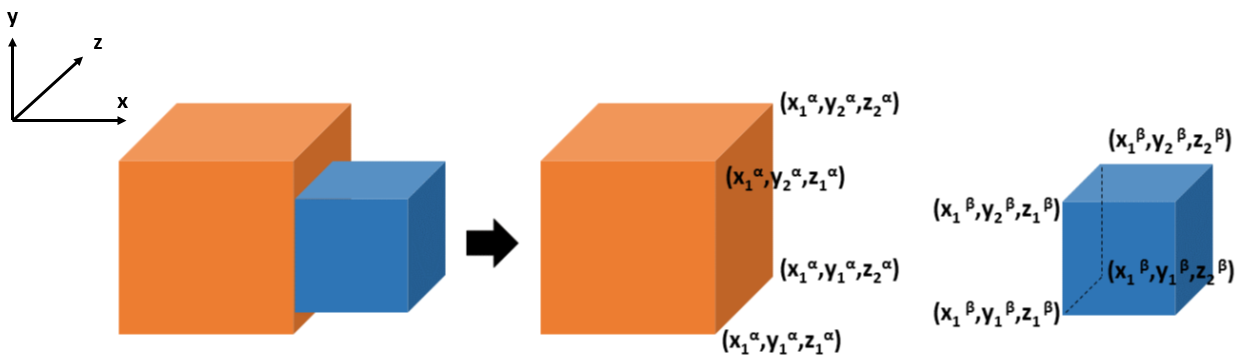}
    \caption{\small Coordinate details of the two cubes that touch each other from side.}
    \label{T_detail}
\end{figure}

For example, in the left second situation of Fig.~\ref{TN}, which has been shown with more details in Fig.\ref{T_detail}, the following condition will lead to a touching relation from a side.

\begin{eqnarray}
&&[x_1^{\beta}=x_1^{\alpha}]  \\ \nonumber
&\land& [(y_1^{\alpha}<y_2^{\beta}<y_2^{\alpha}) \lor (y_1^{\alpha}<y_1^{\beta}<y_2^{\alpha})]  \\ \nonumber
&\land& [(z_1^{\alpha}<z_2^{\beta}<z_2^{\alpha}) \lor (z_1^{\alpha}<z_1^{\beta}<z_2^{\alpha})]
\end{eqnarray}

Moreover, all discussed static and dynamic relations are defined by a set of rules. We start with explaining the rule set for static spatial relations and then proceed to dynamic spatial relations. In general, $x_{min}$, $x_{max}$, $y_{min}$ , $y_{max}$, $z_{min}$ and $z_{max}$ indicate the minimum and maximum values between the points of object cube $\alpha_{i}$ in \textit{x}, \textit{y} and \textit{z} axes, respectively.

Let us define the relation ``Left'', $SSR(\alpha_i, \alpha_j)=\mathrm{\textbf{L}} $ (object $\alpha_i$ is to the left of object $\alpha_j$) if:
\begin{equation}
\begin{split}
x_{max}(\alpha_i)<x_{max}(\alpha_j)
\end{split}
\end{equation}
and the following exception condition holds
\begin{eqnarray}
&&[\neg({y_{min}(\alpha_i)>y_{max}(\alpha_j))]} \nonumber \\
&\land& [\neg({y_{min}(\alpha_j)>y_{max}(\alpha_i))]} \nonumber\\
&\land& [\neg({z_{min}(\alpha_i)>z_{max}(\alpha_j))]} \nonumber\\
&\land& [\neg({z_{min}(\alpha_j)>z_{max}(\alpha_i))]}
\end{eqnarray}

\begin{figure}[!h]
    \centering
    \includegraphics[width=0.75\textwidth]{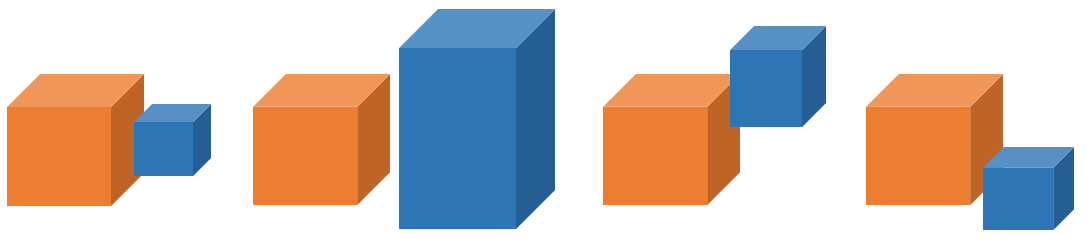}
    \caption{\small Possible states of Left relation between two objects cubes when size and y positions vary.}
    \label{left}
\end{figure}

The exception condition excludes from the relation ``Left'' those cases when two object cubes do not overlap in altitude (y direction) or front/back (z direction). Several examples of objects holding relation $SSR(red, blue)=\mathrm{\textbf{L}} $, when the size and shift in y direction varies, are shown in Fig. \ref{left}.

$SSR(\alpha_i, \alpha_j)=\mathrm{\textbf{R}}$ is defined by $x_{max}(\alpha_i)>x_{min}(\alpha_j) $ and the identical set of exception conditions. The relations \textbf{Ab}, \textbf{Be}, \textbf{F}, \textbf{Ba} are defined in an analogous way. For \textbf{Ab} and \textbf{Be} the emphasis is on the ``\textit{y}'' dimension, while for the \textbf{F}, \textbf{Ba} the emphasis is on the ``\textit{z}'' dimension.

For the relation ``inside'' $SSR(\alpha_i, \alpha_j)=\mathrm{\textbf{In}}$ we use:
\begin{eqnarray}
&&[x_{min}(\alpha_j)\leq x_{min}(\alpha_i)] \land
[x_{max}(\alpha_i)\leq x_{max}(\alpha_j)] \nonumber \\
&\land& [z_{min}(\alpha_j)\leq z_{min}(\alpha_i)] \land
[z_{max}(\alpha_i)\leq z_{max}(\alpha_j)] \nonumber \\
&\land& [y_{min}(\alpha_j)\leq y_{max}(\alpha_i)\leq y_{max}(\alpha_j)]
\end{eqnarray}

The opposite holds for relation \textbf{Sa} (surrounding). For example, if $SSR(\alpha_i, \alpha_j)=\mathrm{\textbf{In}} \implies SSR(\alpha_j, \alpha_i)=\mathrm{\textbf{Sa}}$

In addition of computing spatial relations TNR between two objects based on the above rules, we also check the touching relation between those two objects. This is then used to define several other relations. For example, if one object is above the other object, while they are touching each other, their static relation will be \textbf{To} (top).
\begin{equation}
[SSR(\alpha_i, \alpha_j)=\mathrm{\textbf{Ab}}] \, \land \, [TNR(\alpha_i, \alpha_j)=\mathrm{\textbf{T}}]  \implies [SSR(\alpha_i, \alpha_j)=\mathrm{\textbf{To}}]
\end{equation}

\par
There can be more than one static spatial relations between two object cubes. For example, one object can be both to the left and in back of the other object. However, to fill the eSEC matrix elements we need only one relation per object pair. This problem is solved by definition of a new notion called \textit{shadow}.

\begin{figure}[!t]
    \centering
    \includegraphics[width=1\textwidth]{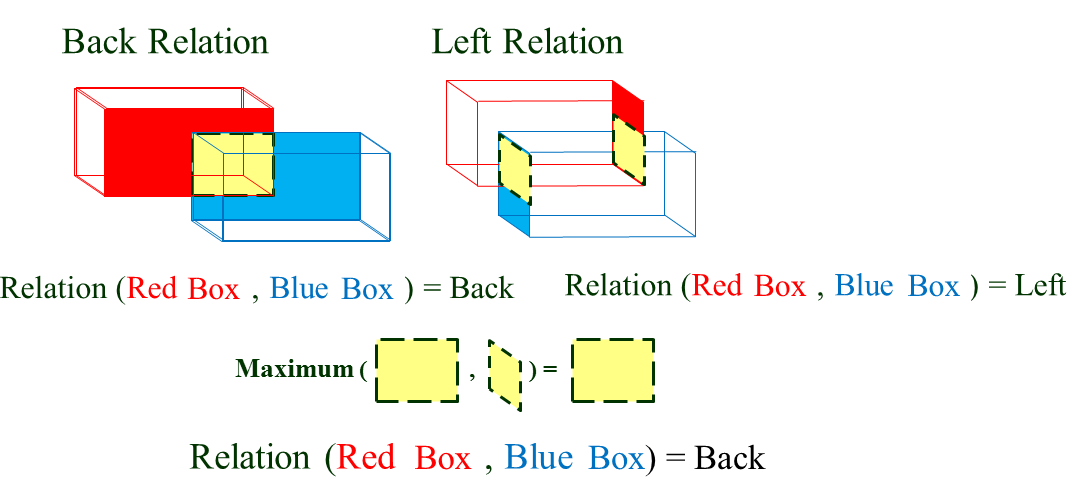}
    \caption{\small Selection of one static spatial relation from several possible relations.}
    \label{multi}
\end{figure}

Each cube has six surfaces. We label them as top, bottom, right, left, front and back based on their positions in our scene coordinate system. Whenever object $\alpha_{i}$ is to the left of object $\alpha_{j}$, one can make a projection from the right surface of object $\alpha_{i}$ onto the left rectangle of object $\alpha_{j}$ and consider only the rectangle intersection area, This area is represented by the newly defined parameter \textit{shadow}.
Suppose $SSR(\alpha_{i},\alpha_{j})=\{R_{1},...,R_{k}\}$ while ${R_{1},...,R_{m}} \in SSR$ and we have calculated  the $shadow(\alpha_{i},\alpha_{j},R)$ for all relations $R$ between the objects $\alpha_{i}$ and $\alpha_{j}$. The relation with the biggest shadow is then selected as the main static relation between the two objects: (Fig.\ref{multi} includes the above description in the image format.)

\begin{eqnarray}
&SSR(\alpha_{i},\alpha_{j})= R_{n}  (1\leq n\leq k),
if: \ nonumber \\
&shadow(\alpha_{i}, \alpha_{j}, R_{n})= max_{1\leq m \leq k}(Shadow(\alpha_{i},\alpha_{j},R_{m}))
\end{eqnarray}

\par Dynamic spatial relations (DSR) are defined as follows. Suppose $O{i}^{f}$ shows the central point of the object cube $\alpha_{i}^{f}$ (object $\alpha_i$ in $f_{th}$ frame); we define $\delta(\alpha_{i}^{f},\alpha_{j}^{f})=||O{i}^{f}-O{j}^{f} ||$ to be a two argument function for measuring the Euclidean distance between the cubes $\alpha_i$ and $\alpha_j$ in $ f_{th} $ frame.
\begin{equation}
  DSR(\alpha_{i}^{f},\alpha_{j}^{f})=\left\{
  \begin{array}{@{}ll@{}}
    GC, & \text{if}\ \delta(\alpha_{i}^{f},\alpha_{j}^{f}) - \delta(\alpha_{i}^{f+\theta},\alpha_{j}^{f+\theta})>\xi \\
    MA, & \text{if}\ \delta(\alpha_{i}^{f+\theta},\alpha_{j}^{f+\theta})-\delta(\alpha_{i}^{f},\alpha_{j}^{f})>\xi
  \end{array}\right.
  \label{DSR}
\end{equation}
For this we use a time window of $ \theta=10 $ frames (image snapshots in VR) in our experiments $(=0.5s)$; the threshold $ \xi $ is kept at 0.1 m:

In the following we defined five conditions \textbf{P1} to \textbf{P5}, which then will be used to characterize the remaining DSRs.

\begin{eqnarray}
&\textbf{P1:}& [TNR(\alpha_{i}^{f},\alpha_{j}^{f})=\textbf{T}]  \land [TNR(\alpha_{i}^{f+\theta},\alpha_{j}^{f+\theta})=\textbf{T}] \nonumber \\
&\textbf{P2:}& [TNR(\alpha_{i}^{f},\alpha_{j}^{f})=\textbf{N}]  \land [TNR(\alpha_{i}^{f+\theta},\alpha_{j}^{f+\theta})=\textbf{N}] \nonumber \\
&\textbf{P3:}& O_{i}^{f} \neq O{i}^{f+\theta} \nonumber\\
&\textbf{P4:}& O_{j}^{f} \neq O{j}^{f+\theta} \nonumber\\
&\textbf{P5:}& \delta(\alpha_{i}^{f+\theta},\alpha_{j}^{f+\theta})-\delta(\alpha_{i}^{f},\alpha_{j}^{f})<\xi
\label{Ps}
\end{eqnarray}

The dynamic relations \textbf{MT}, \textbf{HT}, \textbf{FMT} and \textbf{S}, based on the five conditions above are now defined in the following way:
\begin{equation}
    DSR(\alpha_{i}^{f},\alpha_{j}^{f})=
    \begin{cases}
      MT, & \text{if}\ P1 \land P3 \land P4 \\
      HT, & \text{if}\ P1 \land \neg P3 \land \neg P4\\
      FMT, & \text{if}\ P1 \land (P3 \oplus P4)\\
      S, & \text{if}\ P2 \land P5  \\
    \end{cases}
    \label{dynamics}
  \end{equation}

\subsection*{Similarity Measure between eSECs}
\label{sim_section}

Suppose $\theta_1$ and $\theta_2$ are the names of two actions with eSECs that have $n$ and $m$ columns, respectively.

Instead of writing down a 30-row eSEC each, we can concatenate the corresponding \textbf{TNR}, \textbf{SSR} and \textbf{DSR} of each fundamental object pair into a triple and make a 10-row matrix for $\theta_1 $ and $\theta_2$ with ternary elements (TNR, SSR, DSR) instead:

\newcommand\scalemath[2]{\scalebox{#1}{\mbox{\ensuremath{\displaystyle #2}}}}
\[
\theta_1 =\left(
    \begin{array}{cccccccc}
  (a_{1,1},a_{11,1},a_{21,1})& (a_{1,2},a_{11,2},a_{21,2})& \cdots &(a_{1,n},a_{11,n},a_{21,n})\\
  (a_{2,1},a_{12,1},a_{22,1})& (a_{2,2},a_{12,2},a_{22,2})& \cdots &(a_{2,n},a_{12,n},a_{22,n}) \\
  \vdots  & \vdots & \ddots &  \vdots  \\
  (a_{10,1},a_{20,1},a_{30,1})& (a_{10,2},a_{20,2},a_{30,2}& \cdots &(a_{10,n},a_{20,n},a_{30,n})
    \end{array}
  \right)
\]
\[
\theta_2 =\left(
    \begin{array}{cccccccc}
  (b_{1,1},b_{11,1},b_{21,1})& (b_{1,2},b_{11,2},b_{21,2})& \cdots &(b_{1,n},b_{11,n},b_{21,n})\\
  (b_{2,1},b_{12,1},b_{22,1})& (b_{2,2},b_{12,2},b_{22,2})& \cdots &(b_{2,n},b_{12,n},b_{22,n}) \\
  \vdots  & \vdots & \ddots &  \vdots  \\
  (b_{10,1},b_{20,1},b_{30,1})& (b_{10,2},b_{20,2},b_{30,2}& \cdots &(b_{10,n},b_{20,n},b_{30,n})
    \end{array}
      \right)
\]

Using the elements of both matrices, we define the differences in the three different relation categories $L^{1:3}$ by:
\[
    L^1_{ i,j}=
\begin{cases}
    0,& \text{if } a_{i,j}=b_{i,j}\\
    1,              & \text{otherwise}
\end{cases}
\]
\[
    L^2_{ i,j}=
\begin{cases}
    0,& \text{if } a_{i+10,j}=b_{i+10,j}\\
    1,              & \text{otherwise}
\end{cases}
\]
\[
    L^3_{ i,j}=
\begin{cases}
    0,& \text{if } a_{i+20,j}=b_{i+20,j}\\
    1,              & \text{otherwise}
\end{cases}
\]

where $ 1\leq i \leq10, 1\leq j \leq k$, $k=max(n,m)$

Then we define the compound difference for the three categories in the following way:

\begin{equation}
\label{sim}
\begin{split}
d_{i,j}=\frac{\sqrt{L^1_{i,j}+ L^2_{i,j}+ L^3_{i,j}}}{\sqrt{3}}.
\end{split}
\end{equation}
In case one matrix had more columns than the other matrix. i.e., $m < n$ or vice versa, we repeated the last column of the smaller matrix to match the number of columns of the bigger matrix. This leads to a consistent drop in similarity regardless of which two action are being compared.

Now we define $D$ as the matrix, which holds all compound differences between the elements of the two eSECs.
$$
{D}_{(10,k)} =
 \begin{pmatrix}
  d_{1,1} & d_{1,2} & \cdots & d_{1,k} \\
  d_{2,1} & d_{2,2} & \cdots & d_{2,k} \\
  \vdots  & \vdots  & \ddots & \vdots  \\
  d_{10,1} & d_{10,2} & \cdots & d_{10,k}
 \end{pmatrix}
$$
where $d_{i,j}$ denotes the dissimilarity of $i_{th}$ objects pair at the $j_{th}$ time stamp (column). Then, $D$, which is the total dissimilarity between eSECs of $\theta_1$ and $\theta_2$ is obtained as the average across all elements of matrix $D$.
\begin{equation}
\label{sim_mat}
D_{\theta_1,\theta_2}=\frac{1}{k*10}(\sum_{j=1}^{k}\sum_{i=1}^{10}d_{i,j})
\end{equation}
Accordingly, the \emph{similarity} between these eSECs $Sim_{\theta_1,\theta_2}$, is obtained as:
\begin{equation}
\label{sim_final}
Sim_{\theta_1,\theta_2}=(1- D_{\theta_1,\theta_2})*100\%
\end{equation}

\subsection*{Statistical Data Analysis}
Data were analyzed using RStudio (Version 1.2.5001, RStudio Inc.) and SPSS 26 (IBM, New York, United States).

To inspect the presence of learning effects in the human sample, correlations (Spearman Rho) were calculated for the number of trial (1 - 30) per action and predictive power as well as error rate.

To compare human and machine predictive power, first, a repeated measures ANOVA on predictive power of humans was calculated with action (1 - 10) as within-subject factor. Then, human and machine performance was compared for each action separately using one-sample t-tests. As the machine data do not show variance, their predictive power value was used to compare it to human performance.

To model human action prediction based on eSEC matrices, we calculated the informational gain based on each eSEC column entry. More specifically, based on the eSEC descriptions of all ten actions, we derived a measurement of the amount of information presented in each column (or action step) of each action in comparison to all other actions. Each eSEC column, for a given sub-table (Touching=T, Static=S, Dynamic=D), contains ten coded descriptions of the spatial relations between hand, objects and ground. By stringing the eSEC codes of one column together, each column gets a new single code formally describing the action stage of a sub-table the participant observes at that moment. By taking the frequency of each action step or column-code across all 10 actions, we calculated the likelihood of a specific code in reference to the other actions in this column. So, if all eSEC descriptions are the same for one column, this column-code is assigned a likelihood of “1”. If only one action differs (from the remaining nine actions), it gets a likelihood of 0.1 and the column-code of the differing action receives a likelihood value of 0.9, and so forth. Because not every action has the same number of columns, the lack of eSEC descriptions is also treated as a possible event. That means, if for example seven out of ten actions already have stopped at one point in time, these seven actions would receive a likelihood of 0.7 for this specific column.

We conducted this likelihood assignment procedure for each of the three types of information (TNR, SSR, DSR) separately. Note that the likelihood also gives an estimate of the information about one action that is presented in a column. If the likelihood of an action code is low, only a few or just this single action has this particular action code. So, if this code appears, it powerfully constrains action prediction.

Based on the likelihood p of an action step x, we then calculated \textit{bit rates} to quantify (self-)information I according to Shannon  \cite{Shannon_1948}:

\begin{equation*}
  I(x)=-log_{2}{(p_x)}
\end{equation*}

This transformation into information has two advantages over calculating with likelihoods. Firstly, it is more intuitive because more information is also displayed as a higher value, and secondly, we now were able to derive cumulated information by adding up the information values associated with successive columns. The transformation and cumulation were also done for all three information types separately. Thus, we obtained information values for each action step for each type of information separately. The additivity of the data also made it possible to combine multiple types of information by simply summing up the columns of the sub-tables.

Based on these information values, we modeled human performance. We employed the following models: one based only on TNR, one based only on SSR, one based only on DSR, three models adding two of the three types of information (T+S; T+D; S+D), one model adding all three types of information (T+S+D) and finally one model that ignores the three differing types of information and calculates the self-information based on all eSEC entries independent of the information type (Overall). For each model and for each action separately, a logistic regression was calculated using SPSS26. Each logistic regression included the absolute amount of information per action step according to the respective model, the accumulated information up to each action step, and the interaction term of these absolute and accumulated predictors. The logistic regressions' dependent binary variable was the presence of a response during the respective action step, indicating whether the action was predicted during this action step or not. Since predictors were correlated, models were estimated using the stepwise forward method for variable entry. Note that we did not interpret the coefficients and therefore did not need to regularize the regression model due to coefficient's correlation. Model fits were compared using the BIC (Bayesian-Information-Criterion) \cite {Schwarz_1978}.



\textbf{Author Contributions:}\\
Designing the study and conceived the idea: F.Z., M.T. and F.W.\\
Recording the VR data: F.Z. and S.P.\\
Performing the VR experiment: F.Z.\\
Data analysis: F.Z. and J.P.\\
Interpreting the data: F.Z., J.P., N.E., R.S., M.T. and F.W.\\
Writing the paper: F.Z., J.P. and R.S.\\
Supervision and providing critical revision: F.W., M.T. and R.S.\\

\textbf{Competing interest}\\
The authors declare no competing interests.\\

\end{document}